\documentclass[journal]{IEEEtran}
\ifCLASSINFOpdf
   \usepackage[pdftex]{graphicx}
   \graphicspath{{figures/}}
\else
\fi
\usepackage{amsmath}
\usepackage{amsfonts} 
\newcommand{\specialcell}[2][c]{%
  \begin{tabular}[#1]{@{}c@{}}#2\end{tabular}}

\usepackage[export]{adjustbox}
\usepackage{enumitem}
\usepackage{comment}
\usepackage{booktabs}
\usepackage{color}
\usepackage[table]{xcolor}
\usepackage{arydshln}
\usepackage{float}
\usepackage{multirow}

\usepackage[linesnumbered,ruled]{algorithm2e}
\SetAlFnt{\small}
\SetAlCapFnt{\small}
\SetAlCapNameFnt{\small}

\SetCommentSty{mycommfont}

\begin{document}
%
\title{pysamoo: Surrogate-Assisted Multi-Objective Optimization in Python}

%

\author{Julian~Blank and Kalyanmoy~Deb
\thanks{Julian Blank and Kalyanmoy Deb are at Michigan State University, East Lansing, USA (email: \{blankjul,kdeb\}@msu.edu).}
}

%
%

\markboth{}%
{Shell \MakeLowercase{\textit{et al.}}: Bare Demo of IEEEtran.cls for IEEE Journals}
%



\maketitle

\begin{abstract}
Significant effort has been made to solve computationally expensive optimization problems in the past two decades, and various optimization methods incorporating surrogates into optimization have been proposed.
However, most optimization toolboxes do not consist of ready-to-run algorithms for computationally expensive problems, especially in combination with other key requirements, such as handling multiple conflicting objectives or constraints. Thus, the lack of appropriate software packages has become a bottleneck for solving real-world applications.
The proposed framework, pysamoo, addresses these shortcomings of existing optimization frameworks and provides multiple optimization methods for handling problems involving time-consuming evaluation functions. The framework extends the functionalities of pymoo, a popular and comprehensive toolbox for multi-objective optimization, and incorporates surrogates to support expensive function evaluations. 
The framework is available under the GNU Affero General Public License (AGPL) and is primarily designed for research purposes. For more information about pysamoo, readers are encouraged to visit: anyoptimization.com/projects/pysamoo.
\end{abstract}


%
\IEEEpeerreviewmaketitle

\section{Expensive Optimization}

\IEEEPARstart{M}{any} optimization problems are computationally expensive and require the execution of one or multiple time-consuming functions to evaluate a solution. Expensive Optimization Problems (EOPs) are especially important in practice and are omnipresent in all kinds of research and application areas, for instance~Agriculture~\cite{2019-roy-crop-yield},~Engineering~\cite{2016-yin-crashwrothiness-design}, Health Care~\cite{2016-lucidi-application-health}, or Computer Science~\cite{2019-lu-nsga-net}. Often the expensiveness of the evaluation is caused by the requirement of running a simulation, such as Computational Fluid Dynamic (CFD)~\cite{1995-anderson-cfd}, Finite Element Analysis (FEA)~\cite{1991-szabo-fea}, or processing a large amount of data~\cite{2019-luo-data-driven,2020-wang-data-driven-forest}.
It is worth noting that the majority of simulation-based data-intensive problems are black-box in nature~\cite{2002-olafsson-simopt} and gradient information is not available or is even more time-consuming to derive.
This makes it even more vital to address the time-consuming objective and/or constraint functions as an inherent part of the optimization problem and significantly limits the overall evaluation budget.

\begin{figure*}[t]
    \centering
    \includegraphics[page=2,width=0.8\linewidth,trim=0 2.5cm 1cm 0,clip]{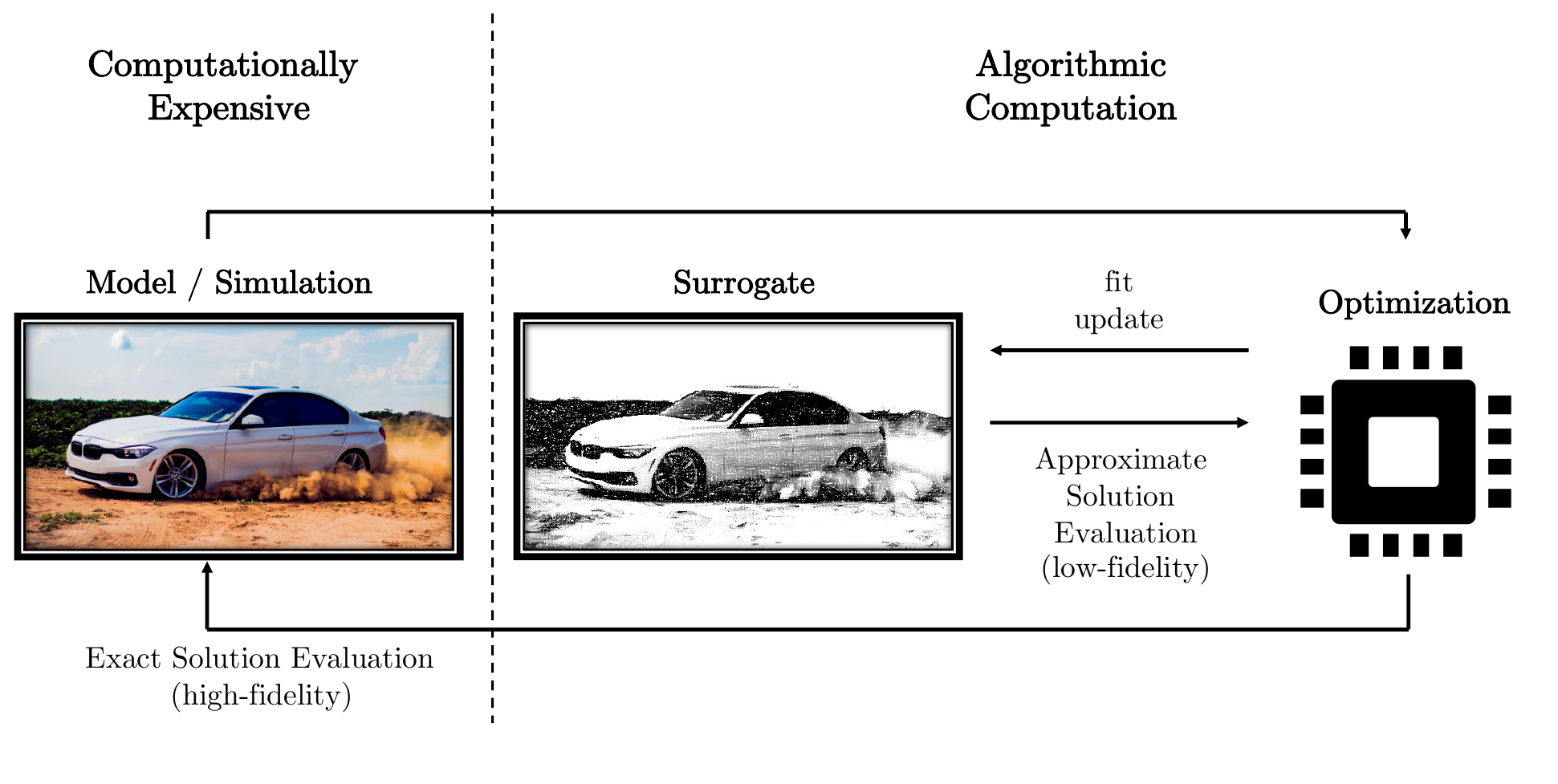}
    \caption{The relation between expensive simulations and optimization.}
    \label{fig:surrogate}
\end{figure*}

The execution of a time-consuming evaluation dominates the algorithm's computational overhead for finding new solutions in each iteration. Thus, an algorithm has more time for carefully selecting new designs than traditional optimization algorithms; however, the evaluation budget is usually only limited to a few hundred evaluations instead of a few thousand evaluations.
A standard method to speed up the convergence of existing methods is using a \emph{surrogate} model (also called metamodel, approximation model, simulation model, data-driven model, response surface), which approximates the time-consuming function. 
Incorporating a surrogate into the optimization process is indicated by adding "surrogate-assisted" or "metamodel-based" to the algorithm's name or description. The incorporation of surrogates into optimization is illustrated in Figure~\ref{fig:surrogate}. 
Commonly, surrogates -- approximation or interpolation models -- are utilized during optimization to improve the convergence behavior. 
First, one shall distinguish between two different types of evaluations: Expensive solution evaluations (ESEs), which require running the computationally expensive evaluation, and approximate solution evaluations (ASEs), which is a computationally inexpensive approximation by the surrogate. 
Where the overall optimization run is limited by $\text{ESE}^{\max}$ function evaluation, function calls of ASEs are only considered as algorithmic overhead. 
The goal of surrogate-assisted optimization is to provide efficient ASEs with the least approximation error possible and to exploit them to minimize ESEs function calls. Thus, the overall goal is to improve the convergence of an optimization algorithm as much as possible in the usually very limited evaluation budget $\text{ESE}^{\max}$.

\begin{table}
     \caption{Terminology of different aspects using surrogates.}
    \centering
    \begin{tabular}{|l|c|c|}
        \toprule
        \textbf{Aspect} & \textbf{Model / Simulation}   & \textbf{Surrogate} \\
        \midrule
        \textbf{Time} &  \specialcell{computationally expensive\\time-consuming} & \specialcell{computationally\\inexpensive } \\[2mm] \hline
        \textbf{Accuracy}   &  high-fidelity & low-fidelity\\ \hline
        \textbf{Evaluation} &  \specialcell{Expensive Solution\\Evaluation (ESE)}  & \specialcell{Approximate Solution\\Evaluation (ASE)}  \\
        \bottomrule
    \end{tabular}
    \label{tab:terminology}
\end{table}

\section{The Framework}

Most researchers rely on open-source packages to conduct their research. Finding an appropriate open-source package can be quite challenging, especially when attempting to solve real-world and application problems.
For optimization, one has to note that most optimization toolboxes do not consist of ready-to-run algorithms for computationally expensive problems.
Moreover, other vital requirements, such as handling multiple conflicting objectives or constraints, are often not supported. 
The proposed framework, pysamoo, addresses these shortcomings of existing optimization frameworks and provides different types of optimization methods targeting time-consuming functions. The framework extends the functionalities of pymoo~\cite{2020-blank-pymoo}, a popular and comprehensive toolbox for multi-objective optimization, and incorporates surrogate support.

The framework includes implementations of PSAF~\cite{2021-blankjul-psaf} and GPSAF~\cite{2022-blankjul-gpsaf}, which are two surrogate incorporation strategies (proposed by the authors of this paper) to generalize model assistance for all different types of metaheuristics.
In the current version of the framework surrogates are incorporated into GA~\cite{1989-goldberg-ga}, DE~\cite{1997-storn-de}, CMAES~\cite{2001-hansen-cmaes},
ISRES~\cite{2005-runarsson-isres}
NSGA-II~\cite{2002-deb-nsga2}, and NSGA-III~\cite{2014-deb-nsga3-part1, 2014-deb-nsga3-part2,2019-blank-nsga3-norm} and others.
The framework is available under the GNU Affero General Public License (AGPL) and is primarily designed for research purposes.
For further information on the constituent PSAF and GPSAF algorithms which are in the core of pysamoo, refer to references \cite{2021-blankjul-psaf} and \cite{2022-blankjul-gpsaf}, respectively.

\ifCLASSOPTIONcaptionsoff
  \newpage
\fi



\bibliographystyle{IEEEtran}
\bibliography{IEEEabrv,references}

\begin{thebibliography}{10}
\providecommand{\url}[1]{#1}
\csname url@samestyle\endcsname
\providecommand{\newblock}{\relax}
\providecommand{\bibinfo}[2]{#2}
\providecommand{\BIBentrySTDinterwordspacing}{\spaceskip=0pt\relax}
\providecommand{\BIBentryALTinterwordstretchfactor}{4}
\providecommand{\BIBentryALTinterwordspacing}{\spaceskip=\fontdimen2\font plus
\BIBentryALTinterwordstretchfactor\fontdimen3\font minus
  \fontdimen4\font\relax}
\providecommand{\BIBforeignlanguage}[2]{{%
\expandafter\ifx\csname l@#1\endcsname\relax
\typeout{** WARNING: IEEEtran.bst: No hyphenation pattern has been}%
\typeout{** loaded for the language `#1'. Using the pattern for}%
\typeout{** the default language instead.}%
\else
\language=\csname l@#1\endcsname
\fi
#2}}
\providecommand{\BIBdecl}{\relax}
\BIBdecl

\bibitem{2019-roy-crop-yield}
P.~C. Roy, A.~Guber, M.~Abouali, A.~P. Nejadhashemi, K.~Deb, and A.~J.~M.
  Smucker, ``Simulation {Optimization} of {Water} {Usage} and {Crop} {Yield}
  {Using} {Precision} {Irrigation},'' in \emph{Evolutionary {Multi}-{Criterion}
  {Optimization}}, K.~Deb, E.~Goodman, C.~A. Coello~Coello, K.~Klamroth,
  K.~Miettinen, S.~Mostaghim, and P.~Reed, Eds.\hskip 1em plus 0.5em minus
  0.4em\relax Cham: Springer International Publishing, 2019, pp. 695--706.

\bibitem{2016-yin-crashwrothiness-design}
H.~Yin, H.~Fang, G.~Wen, Q.~Wang, and Y.~Xiao, ``An adaptive {RBF}-based
  multi-objective optimization method for crashworthiness design of
  functionally graded multi-cell tube,'' \emph{Structural and Multidisciplinary
  Optimization}, vol.~53, no.~1, pp. 129--144, 2016.

\bibitem{2016-lucidi-application-health}
S.~Lucidi, M.~Maurici, L.~Paulon, F.~Rinaldi, and M.~Roma, ``A simulation-based
  multiobjective optimization approach for health care service management,''
  \emph{IEEE Transactions on Automation Science and Engineering}, vol.~13,
  no.~4, pp. 1480--1491, 2016.

\bibitem{2019-lu-nsga-net}
\BIBentryALTinterwordspacing
Z.~Lu, I.~Whalen, V.~Boddeti, Y.~Dhebar, K.~Deb, E.~Goodman, and W.~Banzhaf,
  ``{NSGA}-{Net}: {Neural} architecture search using multi-objective genetic
  algorithm,'' in \emph{Proceedings of the genetic and evolutionary computation
  conference}, ser. {GECCO} ’19.\hskip 1em plus 0.5em minus 0.4em\relax New
  York, NY, USA: Association for Computing Machinery, 2019, pp. 419--427.
  [Online]. Available: \url{https://doi.org/10.1145/3321707.3321729}
\BIBentrySTDinterwordspacing

\bibitem{1995-anderson-cfd}
J.~D. Anderson and J.~Wendt, \emph{Computational fluid dynamics}.\hskip 1em
  plus 0.5em minus 0.4em\relax Springer, 1995, vol. 206.

\bibitem{1991-szabo-fea}
B.~Szabó and I.~Babuška, \emph{Finite element analysis}.\hskip 1em plus 0.5em
  minus 0.4em\relax John Wiley \& Sons, 1991.

\bibitem{2019-luo-data-driven}
W.~Luo, R.~Yi, B.~Yang, and P.~Xu, ``Surrogate-assisted evolutionary framework
  for data-driven dynamic optimization,'' \emph{IEEE Transactions on Emerging
  Topics in Computational Intelligence}, vol.~3, no.~2, pp. 137--150, 2019.

\bibitem{2020-wang-data-driven-forest}
H.~Wang and Y.~Jin, ``A random forest-assisted evolutionary algorithm for
  data-driven constrained multiobjective combinatorial optimization of trauma
  systems,'' \emph{IEEE Transactions on Cybernetics}, vol.~50, no.~2, pp.
  536--549, 2020.

\bibitem{2002-olafsson-simopt}
S.~Olafsson and J.~Kim, ``Simulation optimization,'' in \emph{Proceedings of
  the winter simulation conference}, vol.~1, 2002, pp. 79--84.

\bibitem{2020-blank-pymoo}
J.~Blank and K.~Deb, ``pymoo: {Multi}-objective {Optimization} in {Python},''
  \emph{IEEE Access}, vol.~8, pp. 89\,497--89\,509, 2020.

\bibitem{2021-blankjul-psaf}
\BIBentryALTinterwordspacing
------, ``{PSAF}: {A} {Probabilistic} {Surrogate}-{Assisted} {Framework} for
  {Single}-{Objective} {Optimization},'' in \emph{{GECCO} '21: {Proceedings} of
  the genetic and evolutionary computation conference companion}.\hskip 1em
  plus 0.5em minus 0.4em\relax New York, NY, USA: ACM, 2021, place: New York,
  NY, USA. [Online]. Available: \url{https://doi.org/10.1145/3449639.3459297}
\BIBentrySTDinterwordspacing

\bibitem{2022-blankjul-gpsaf}
------, ``{GPSAF}: {A} {Generalized} {Probabilistic} {Surrogate}-{Assisted}
  {Framework} for {Constrained} {Single}- and {Multi}-objective
  {Optimization},'' \emph{IEEE Transactions on Evolutionary Computation}.

\bibitem{1989-goldberg-ga}
D.~E. Goldberg, \emph{Genetic algorithms in search, optimization and machine
  learning}, 1st~ed.\hskip 1em plus 0.5em minus 0.4em\relax USA: Addison-Wesley
  Longman Publishing Co., Inc., 1989.

\bibitem{1997-storn-de}
R.~Storn and K.~Price, ``Differential {Evolution} –{A} {Simple} and
  {Efficient} {Heuristic} for global {Optimization} over {Continuous}
  {Spaces},'' \emph{Journal of Global Optimization}, no.~4, pp. 341--359, Dec.
  1997.

\bibitem{2001-hansen-cmaes}
\BIBentryALTinterwordspacing
N.~Hansen and A.~Ostermeier, ``Completely derandomized self-adaptation in
  evolution strategies,'' \emph{Evolutionary Computation}, vol.~9, no.~2, pp.
  159--195, Jun. 2001. [Online]. Available:
  \url{http://dx.doi.org/10.1162/106365601750190398}
\BIBentrySTDinterwordspacing

\bibitem{2005-runarsson-isres}
T.~Runarsson and X.~Yao, ``Search biases in constrained evolutionary
  optimization,'' \emph{IEEE Transactions on Systems, Man, and Cybernetics,
  Part C (Applications and Reviews)}, vol.~35, no.~2, pp. 233--243, 2005.

\bibitem{2002-deb-nsga2}
\BIBentryALTinterwordspacing
K.~Deb, A.~Pratap, S.~Agarwal, and T.~Meyarivan, ``A fast and elitist
  multiobjective genetic algorithm: {NSGA}-{II},'' \emph{Trans. Evol. Comp},
  vol.~6, no.~2, pp. 182--197, Apr. 2002. [Online]. Available:
  \url{https://doi.org/10.1109/4235.996017}
\BIBentrySTDinterwordspacing

\bibitem{2014-deb-nsga3-part1}
K.~Deb and H.~Jain, ``\BIBforeignlanguage{English (US)}{An evolutionary
  many-objective optimization algorithm using reference-point-based
  nondominated sorting approach, {Part} {I}: {Solving} problems with box
  constraints},'' \emph{\BIBforeignlanguage{English (US)}{IEEE Transactions on
  Evolutionary Computation}}, vol.~18, no.~4, pp. 577--601, 2014.

\bibitem{2014-deb-nsga3-part2}
H.~Jain and K.~Deb, ``An evolutionary many-objective optimization algorithm
  using reference-point based nondominated sorting approach, part {II}:
  {Handling} constraints and extending to an adaptive approach,'' \emph{IEEE
  Transactions on Evolutionary Computation}, vol.~18, no.~4, pp. 602--622, Aug.
  2014.

\bibitem{2019-blank-nsga3-norm}
J.~Blank, K.~Deb, and P.~Roy, ``Investigating the normalization procedure of
  {NSGA}-{III},'' in \emph{Evolutionary multi-criterion optimization}, K.~Deb,
  E.~Goodman, C.~A. Coello~Coello, K.~Klamroth, K.~Miettinen, S.~Mostaghim, and
  P.~Reed, Eds.\hskip 1em plus 0.5em minus 0.4em\relax Springer International
  Publishing, 2019, pp. 229--240, place: Cham.

\end{thebibliography}
\end{document}